\documentclass[conference]{IEEEtran}
\IEEEoverridecommandlockouts
\usepackage{cite}
\usepackage{amsmath,amssymb,amsfonts}
\usepackage{algorithm}
\usepackage{algorithmic}
\usepackage{amssymb}
\usepackage{amsmath}
\usepackage{multirow}
\usepackage{booktabs}
\usepackage{pifont}
\usepackage{graphicx}
\usepackage{textcomp}
\usepackage{xcolor}
\usepackage{subfigure}
\usepackage{caption}
\usepackage{pgffor}

\def\BibTeX{{\rm B\kern-.05em{\sc i\kern-.025em b}\kern-.08em
    T\kern-.1667em\lower.7ex\hbox{E}\kern-.125emX}}

\usepackage{fancyhdr}

\begin{document}

\title{JointDistill: Adaptive Multi-Task Distillation for Joint Depth Estimation and Scene Segmentation}

\author{Tiancong Cheng$^{1}$, Ying Zhang$^{1,*}$ \thanks{\vspace{-0.22in}\newline \rule{0.45\linewidth}{0.1pt} \newline * Corresponding authors, Email: \{izhangying, zhiwenyu\}@nwpu.edu.cn \newline  © 2025 IEEE. Personal use of this material is permitted. Permission from IEEE must be obtained for all other uses, in any current or future media, including reprinting/republishing this material for advertising or promotional purposes, creating new collective works, for resale or redistribution to servers or lists, or reuse of any copyrighted component of this work in other works.}, Yuxuan Liang$^{2}$, Roger Zimmermann$^{3}$, Zhiwen Yu$^{1,4,*}$, Bin Guo$^{1}$ \\
\textit{$^{1}$Northwestern Polytechnical University}\\
\textit{$^{2}$Hong Kong University of Science and Technology (Guangzhou)}\\
\textit{$^{3}$National University of Singapore}\\
\textit{$^{4}$Harbin Engineering University}}

\maketitle

\begin{abstract}

Depth estimation and scene segmentation are two important tasks in intelligent transportation systems. A joint modeling of these two tasks will reduce the requirement for both the storage and training efforts. This work explores how the multi-task distillation could be used to improve such unified modeling. While existing solutions transfer multiple teachers' knowledge in a static way, we propose a self-adaptive distillation method that can dynamically adjust the knowledge amount from each teacher according to the student's current learning ability. Furthermore, as multiple teachers exist, the student's gradient update direction in the distillation is more prone to be erroneous where knowledge forgetting may occur. To avoid this, we propose a knowledge trajectory to record the most essential information that a model has learnt in the past, based on which a trajectory-based distillation loss is designed to guide the student to follow the learning curve similarly in a cost-effective way. We evaluate our method on multiple benchmarking datasets including Cityscapes and NYU-v2. Compared to the state-of-the-art solutions, our method achieves a clearly improvement. The code is provided in the supplementary materials.

\end{abstract}


\section{Introduction}

Depth estimation and scene segmentation are two important tasks in intelligent transportation systems~\cite{muhammad2022vision}. The early studies usually treat these two tasks as separate models~\cite{hazirbas2017fusenet,palafox2019semanticdepth,natan2021semantic}, which requires heavy storage as well as separate training efforts. Recent approaches have begun adopting multi-task learning (MTL) to achieve more compact models, with the majority of parameters shared across tasks. Typical multi-task solutions include on e.g., developing new architectures~\cite{ye2022inverted,gurulingan2023multi}, enhancing parameters~\cite{misra2016cross} or structural information sharing across tasks~\cite{yang2023contrastive}, optimizing updating directions~\cite{liu2021conflict,liu2024famo} and keeping gradient balance~\cite{chen2018gradnorm,yu2020gradient}. The multi-task model can be used directly, or served as an image-encoder and combined with other modalities for further improvement~\cite{liang2022effective,liu2023hierarchical}. However, the performance of most MTL solutions degrades heavily compared to the single-task models and thus knowledge distillation (KD) methods are a popular method for accuracy compensation.

\begin{figure}[!hbpt]
\centering
\includegraphics[width=3.5in]{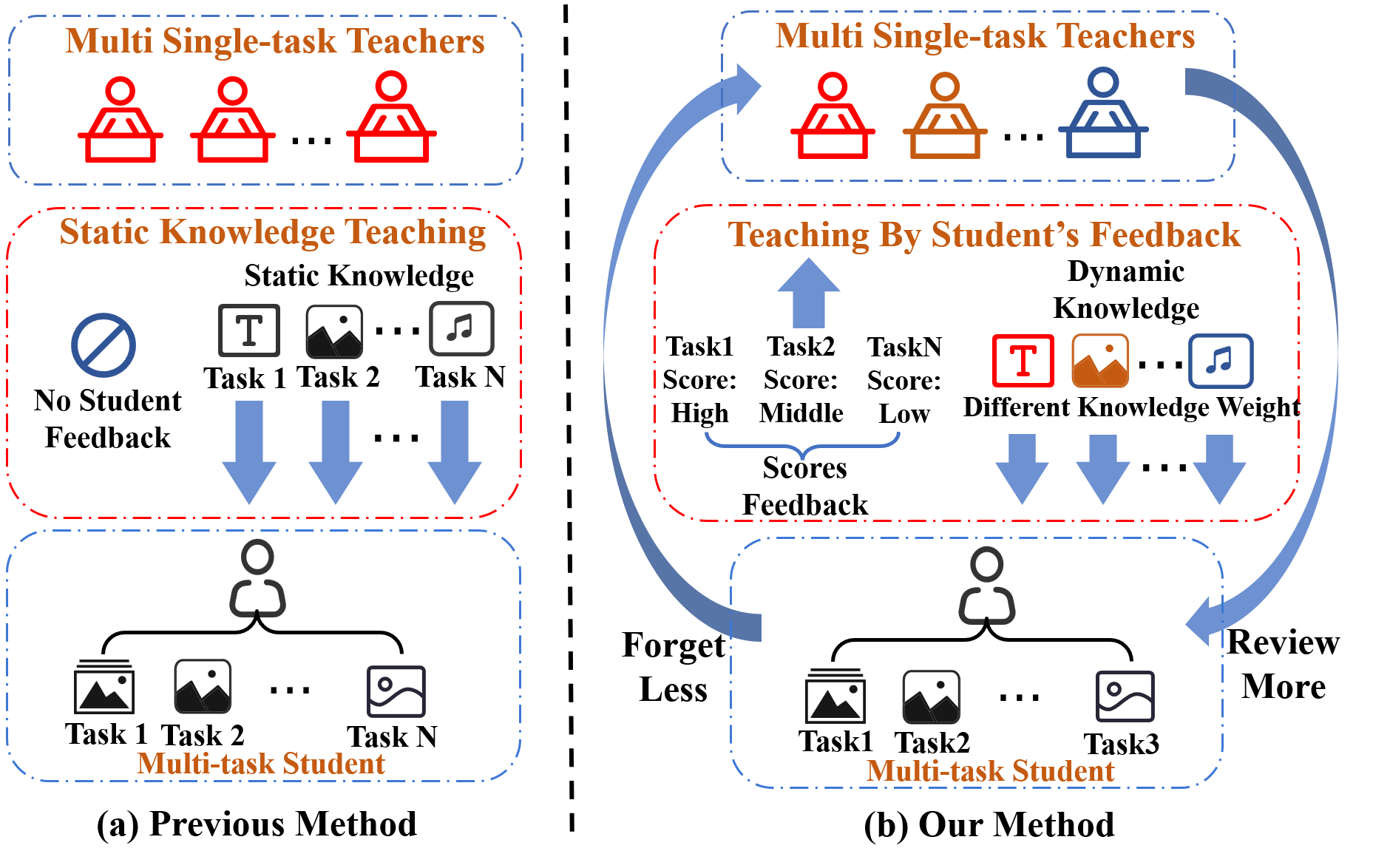} 
\caption{\small Compared to existing static distillation method, we adjust the amount of knowledge to be transferred from each single-task teacher in a dynamic manner based on the student's feedback. Such feedback reflects the true ability of the student on each task so that we could balance the multiple knowledge adaptively.}
\vspace{-3pt}
\label{fig_comparision}
\end{figure}

While most distillation solutions are proposed for a single task, when extending to a multi-task setting, the very first challenge is how to balance the multiple single-task teachers. A common solution is to assign each teacher a weight so that the joint knowledge can be integrated and transferred to the student accordingly~\cite{yun2023achievement,qian2022switchable,xu2023multi}. However, such fixed-weight solutions ignore the dynamic nature of the learning procedure. Specifically, during the training procedure, the model ability might not be the same all the time and therefore we require a self-adaptive mechanism to capture such dynamics. The second challenge is, while multiple teachers exist, the selection of gradient update direction is more prone to be erroneous~\cite{yang2024learning,song2023ecotta} which might trigger the knowledge forgetting problem. Previous methods typically store a large number of early parameters~\cite{niu2022efficient,song2023ecotta} or data~\cite{zancato2023train,wang2022continual}, requiring significant storage overhead. Therefore, how to provide a low-cost mechanism is essential. 

To overcome the above two challenges, we propose a self-adaptive multi-task distillation approach for joint scene segmentation and depth estimation. Specifically, we first propose a feedback-based multi-teacher weight adjustment strategy to dynamically change the knowledge transferring weight from each teacher model (a scene segmentation teacher and a depth estimation teacher in our case). An unseen validation dataset is used to assess the student's current performance for each task which determines the corresponding teacher's weight in the next training period. Secondly,  we propose a knowledge trajectory to record the most significant knowledge of a teacher model that has been learnt in the past. Then the student follows the trajectories of multiple teachers to ensure a similar learning.  Compared to existing solutions, the trajectory is of much lower storage and computation cost as only the essential information is memorized rather than everything. In summary, our contributions are outlined as follows:
\begin{itemize}
\item To the best of our knowledge, it is the first self-adaptive multi-teacher knowledge distillation strategy to jointly solve the scene segmentation and depth estimation tasks. Compared to existing static-weight distillations, we capture the dynamic changes of a student's ability temporally and result a performance improvement. 
\item We introduce a novel knowledge trajectory to record the most important information a model learnt in history and require the student to follow the path based on a the proposed trajectory loss. This design relieves the knowledge forgetting meanwhile reduces the computation and storage requirement from other methods. 
\item Extensive experiments are carried out on two benchmark transportation datasets NYU-V2 and Cityscapes and we observed a 2.83\% and 9.04\% average task improvement, respectively. The multi-task student could be used alone for both tasks, or combined with other modalities for a multi-modal enhancement. 
\end{itemize}

\section{Related Works}

\vspace{3pt}\noindent\textbf{Multi-task Learning Methods}
Existing multi-task learning methods can be categorized into three primary approaches: structural enhancements, consistency-based learning, and balance optimization-based learning. The structural enhancement approaches include the use of trainable parameters~\cite{misra2016cross}, multi-scale information fusion decoder~\cite{ye2022inverted}, and fusion strategy for multi-network modules~\cite{gurulingan2023multi}. Consistency-based methods strive to discern structural similarities between the outputs of different branches in MTL, enabling branches to benefit from each other's consistent structures~\cite{zhang2023faceliveplus}. Multi-task balance optimization can be categorized into loss-based~\cite{liu2019end} and gradient-based~\cite{chen2018gradnorm,cheng2025compmtl} approaches. Additionally, optimization methods based on Lagrangian dual optimization, which aims to maximize the optimal update gradients of each branch, have been extensively studied~\cite{liu2021conflict,liu2024famo}. Meanwhile, feedback-based multi-task collaborative patterns are also crucial in autonomous driving scenarios~\cite{wang2025finghv,wang2024hmos}.

\vspace{3pt}\noindent\textbf{Knowledge Distillation Methods}
Recent methodologies in knowledge distillation can be classified into curriculum learning-based approaches~\cite{li2023curriculum}, attention-based feature aligning~\cite{chen2021cross}, and optimal distillation parameter selection using techniques such as monte carlo tree search~\cite{li2023automated} and evolutionary algorithms~\cite{li2023kd}. While these methods have significantly enhanced the dynamic improvement of student model learning from static teachers in various domains, few have considered dynamically adjusting the teacher's knowledge based on the student model's period performance.

\vspace{3pt}\noindent\textbf{Knowledge Distillation for Multi-task Learning}
Distillations have been implemented in multi-task models via both online~\cite{jacob2023online} and offline~\cite{luo2020collaboration,cheng2023multi} approaches. 
Jacob et al. proposed determining the training weights for multi-task learning based on the correlation between loss variation in single-task and multi-task branches~\cite{jacob2023online}. Xu et al. introduced a novel approach employing low-sensitivity CS divergence alongside intra-class and inter-class losses for distillation~\cite{xu2023multi}. Yun et al. suggested a multi-branch weight determination strategy predicated on the accuracy alignment between teacher and student models~\cite{yun2023achievement}. In these methods, the imbalance resulting from varying task difficulties poses a challenge to the effective application of existing distillation strategies. Among these studies, although task-specific weights are carefully selected, their values are fixed all the time and prevent the student from matching their changing abilities.

The loss function for these multi-task KD methods can be summarized:
\begin{equation}
    L= \sum_{i=1}^{N}  L_{Task}^i+\omega^iL_{KD}^i,
\end{equation} where $L_{Task}^i$ and $L_{KD}^i$ represent $i^{th}$ task-specific loss and KD loss. Typically, KD weights $\omega^i$ must be manually set based on experience, with hyperparameter tuning complexity increasing as the number of tasks grows. We propose an innovative dynamic multi-teacher distillation method based on student's feedback, where knowledge transfer weights $\omega^i$ are dynamically adjusted based on the student's generalization performance during training, rather than through repeated fine-tuning processes.

\begin{figure*}[!hbpt]
\centering
\includegraphics[width=5.5in]{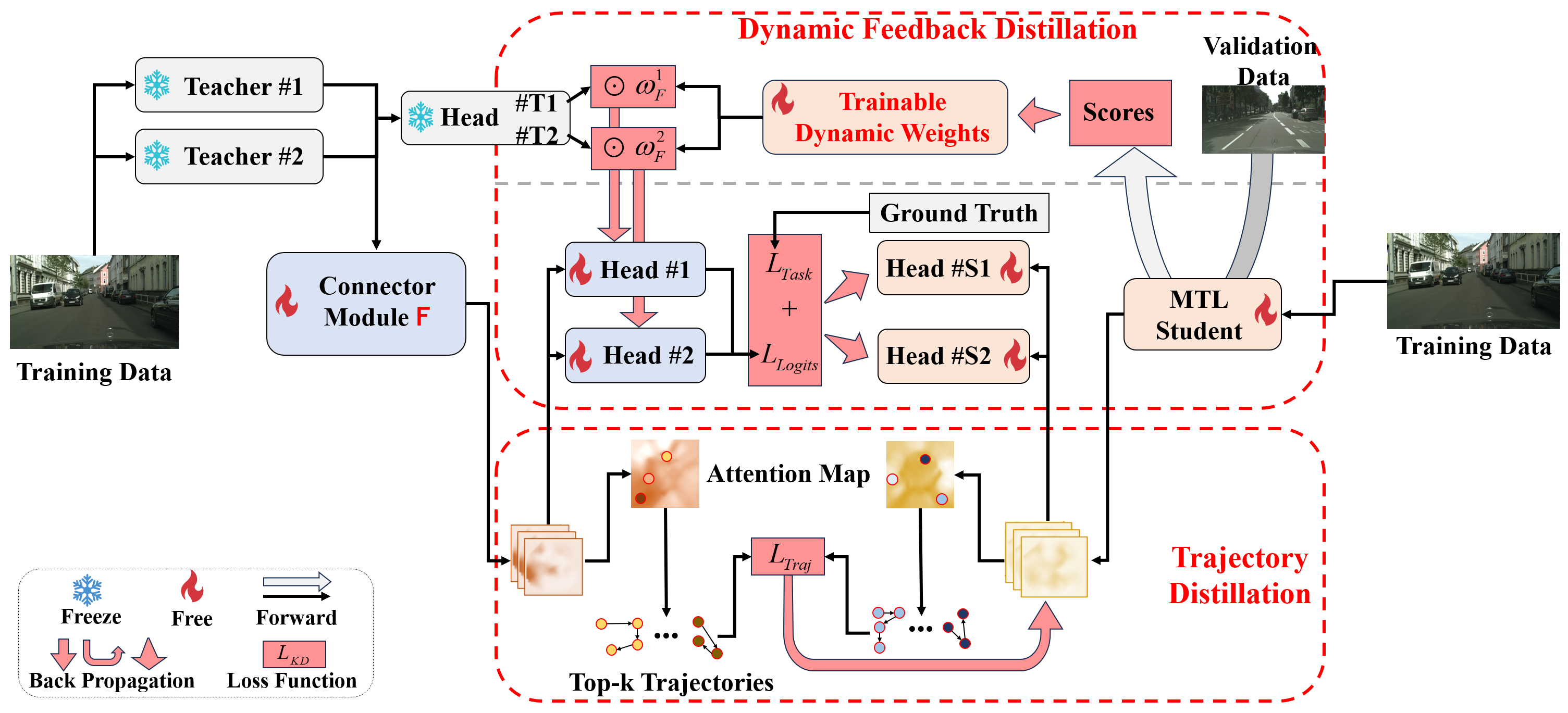}
\caption{\small Illustration of our proposed adaptive multi-task knowledge distillation method. The knowledge from multiple single-task teachers (two teachers in this application, one for segmentation and the other for depth estimation) is integrated and transferred to a multi-task student in a dynamic manner based on the periodical feedback on a separate validation dataset. Meanwhile a novel trajectory distillation is proposed to regulate the student to follow the most essential knowledge of multiple task-specific teachers with lower training cost. }
\label{fig_method}
\vspace{-9pt}
\end{figure*}

\section{Proposed Method}

\subsection{Formulation and Overview}\label{sec: overview} Assume we have $N$ tasks, our goal is to train a multi-task student model $S$ with the best average task performance. The student $S$ is composed of a shared backbone $\mathcal{B}$ and $N$ task-specific heads $\{h_1,h_2,...,h_N\}$. Given a data sample (an RGB image in our case), the output feature of the student's shared backbone is denoted as $f_S$, and the output from $i^{th}$ task-specific head is $p_S^i$. If each task has a well-trained teacher model $T$, our goal is to distill a multi-task student model $S$ with the assistance $N$ single-task teachers $\{T_1, T_2,..., T_N\}$. To connect the multi-teacher model to the student, as illustrated in Fig.~\ref{fig_method}, there usually exists a connector module (denoted by $F$ and $\{h_F^i\}_{i=1}^N$). This module first fuses the features $\{f_{T_i}\}_{i=1}^N$ from $N$ teachers by a feature fusion module $F$ and then transfers the fused feature $f_F$ to the $N$ task-specific heads $h_F^i$. During the fuse operation, the training of the connector module is integrated with dynamic weights $\{\omega_F^i\}_{i=1}^N$. We denote the output of $i^{th}$ connector head as $p_F^i$.

To obtain the student model that can jointly perform well on depth estimation and scene segmentation, we propose an adaptive distillation method and the overview is illustrated in Fig.~\ref{fig_method}. The method is composed of two key modules: 1) A feedback-based multi-teacher distillation aims to dynamically adjust the amount of knowledge from multiple teachers (i.e. the connector heads training weight $\{\omega_F^i|i=1\sim~N\}$ ) during the training, which is determined by the student's feedback on a separate validation dataset;  2) A knowledge trajectory-based distillation aims to record the essential knowledge of the teachers in the historical periods, which guides the student to follow a similar learning curve.   

\subsection{Feedback-based multi-teacher adaptive distillation} 
In this section, we illustrate how to dynamically balance knowledge from multiple teachers. Inspired by human teaching, a good teacher usually adjusts their teaching pace based on student feedback. For example, a positive feedback is received if a student has a good score in an exam, and then the teacher tries to deliver more knowledge in the future. Similarly, we set a separate validation dataset and evaluate the student's performance periodically. We will increase the teacher-specific knowledge transfer in the next training period if with a good performance and decrease otherwise.   

\vspace{3pt}\noindent\textbf{Task-specific feedback estimation.} We divide the training into multiple timeframes. At the $t^{th}$ time frame and for the $i^{th}$ task, we calculate the student performance as $r_S^{i,t}$. If we denote the initial performance of the $i^{th}$ single-task teacher on $\mathbb{D}_{val}$ as $r_{T}^{i,0}$, then the ratio between these two values is a task-specific feedback score: 

\begin{equation}
    a_S^{i,t} = \frac{r_S^{i,t}}{r_{T}^{i,0}}
\label{eqn_achieve}
\end{equation} A higher score indicates that the student is progressing well and a larger weight could be assigned in the next time-frame of that teacher. 

\vspace{3pt}\noindent\textbf{Self-adaptive knowledge adjustment.} Based on the feedback, the weight of that teacher is updated using the formula below:
\begin{equation}
    \omega_F^{i,t+1} =  \omega_F^{i,t} - \beta \omega_F^{i,t} \cdot \nabla_{\omega }\left \{  \left | \omega_F^{i,t}-\overline{\omega_F^{t}} \cdot \left(\frac{a_S^{i,t}}{\overline{a_S^{t}}}\right )^\alpha     \right |_1 \right \}
    \label{equ_omega}
\end{equation} where $\omega_F^{i,t}$ ($\omega_F^{i,t} \propto a_S^{i,t}$) is the dynamic weights, $\overline{a_S^{t}},\overline{\omega_F^{t}}$ denote the average scores among $N$ tasks, $\beta$ represents update step and $\alpha$ is the hyperparameter. We adopt $\omega_F^{i,t}$ to update the connector heads $\{h_F^i\}_{i=1}^N$ in this work. Once the weight is updated for each teacher, as depicted on the middle of the figure~\ref{fig_method}, the training of multi-task connector heads $h_F^i$ will be balanced integrated together and transferred to the student task-specific head accordingly. In this work, the output logits of connector module is denoted as below: 
\begin{equation}
    p_F^i=w_F^i\circ h_F^i\circ \underset{\times 2}{\underbrace{Relu \circ BN \circ Conv}}   \circ \underset{i\epsilon 1,\cdots ,N}{concat} \left \{ f_T^i \right \} .
\label{eqn_ffm}
\end{equation}  Note that, this updated logits is served as new soft-targets for the student to learn and the overall logits loss is converted to the following version:
\begin{equation}
    L_{logits}=\sum_{i=1}^{N} KL(p_S^i \parallel p_F^i).
\label{eqn_stu_kd} 
\end{equation} where $p_S, p_F^i$ are the logits of the $i^{th}$ task-specific head from the student and the connector, respectively. $KL(\cdot)$ represents the Kullback-Leibler divergence, commonly used to measure the distillation loss.

\begin{figure}[!t]
\centering
\includegraphics[width=3.in]{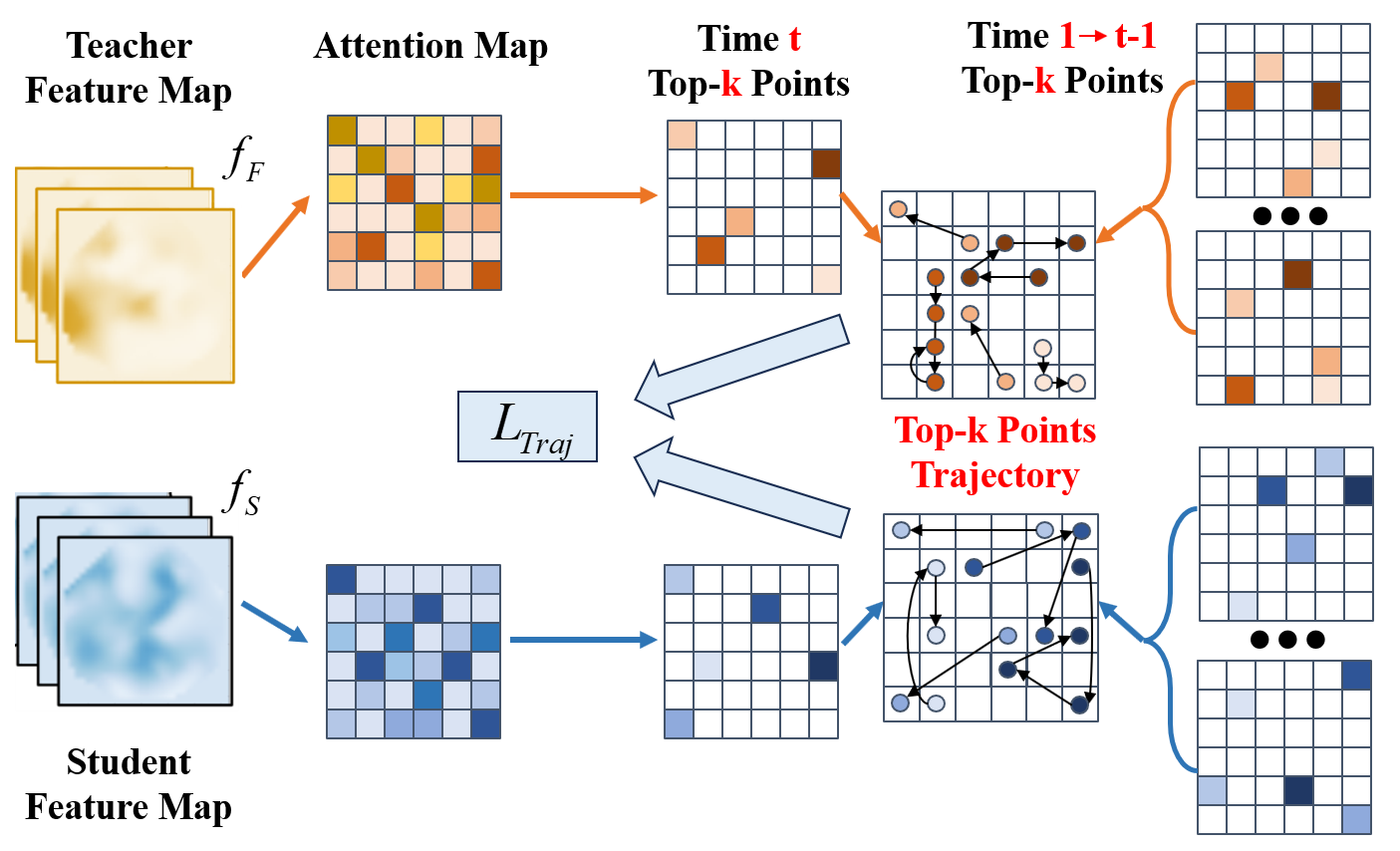} 
\caption{\small In our proposed trajectory distillation, we design a knowledge trajectory to capture the most essential information of the teacher model over the most recent $t$ time-periods, and require the student to follow the top-k trajectories to avoid knowledge forgetting problem.}
\vspace{-9pt}
\label{fig_traj}
\end{figure}

\subsection{Trajectory-based Knowledge distillation}
During the distillation, the student's gradient update direction is influenced by the multiple teachers. Therefore, compared to single-distillation setting, the student is more prone to be erroneous~\cite{yu2020gradient} which might bring in the knowledge forgetting problem~\cite{song2023ecotta,yang2024learning}. To avoid this problem, unlike previous methods by storing a large number of early parameters~\cite{yang2024learning} or data~\cite{song2023ecotta}, we only record the most important knowledge along the timeline which forms a \textit{knowledge trajectory}. The student is enforced to follow this trajectory leveraging a novel trajectory distillation loss.

\vspace{3pt}\noindent\textbf{Knowledge trajectory construction}
To record the most significant information during each training iteration, we first convert the features into the 2D attention maps~\cite{komodakis2017paying} and then extract its essential points. As shown in figure~\ref{fig_traj}, at the $t^{th}$ time-frame, the essential point ($e_t$) is defined as pixel location with the highest values of the 2D attention map. Accordingly, we define the knowledge trajectory as the consecutive essential points along $t$ consecutive time-frames as below: 
\begin{equation}
    J = [e_1, e_2, ..., e_t]
\end{equation}
To ensure sufficient essential knowledge could be included, in stead of using a single trajectory, we further select $K$ trajectories by extracting the top-$k$ essential points in each time-frame:
\begin{equation}
    \mathbb{J} = \{J^1, J^2, ..., J^k\}
\end{equation} where $J^j=[e_1^j, e_2^j, ..., e_t^j]$ is the top-j trajectory and $e^j_t$ is the location with the $j^{th}$ highest values of the 2D attention map at $t^{th}$ time-frame.

\vspace{3pt}\noindent\textbf{Trajectory distillation loss}
Based on the trajectory and to avoid the knowledge forgetting, we require the multi-task student to mimic the top-$K$ trajectores of the multi-task teacher (the connector module, which fuses $N$ single-task teachers) based on our proposed trajectory distillation loss as below:
\begin{equation}
   L_{Traj}= \sum_{j=1}^K Dist(J_{F}^j,J_S^j),
\label{eqn_traj_kd}
\end{equation} where $N$ is the total number of teachers, $J_F\in \mathbb{J}_F$, $J_S\in \mathbb{J}_S$ and $Dist(\cdot)$ is a distance measurement and we adopt the L1 norm in this work. 


\subsection{Training and loss}
Given multiple pretrained teachers, we simultaneously train two parts: the multi-task student and the connector module. What below describes the loss for each part. 

\vspace{3pt}\noindent\textbf{Loss for the student} Based on all above, for the multi-task student model, its final loss is summarized as below:
\begin{equation}
    L_S=\sum_{i=1}^{N} L_{Task}^i+L_{Logits}+\lambda L_{Traj},
\label{eqn_stu}
\end{equation} where $L_{Task}^i$ is the $i^{th}$ task loss (e.g., cross entropy for segmentation and L1 norm for depth estimation).

\vspace{3pt}\noindent\textbf{Loss for connector}
Denote the connector module as $F$, its distillation loss is derived as below: 
\begin{equation}
    L_F=\sum_{i}^{N} \omega_F^i\cdot  KL(p_F^i \parallel p_T^i).
\label{eqn_lt_kd}
\end{equation} In the equation, $KL(\cdot )$ is Kullback-Leibler divergence. $p_F^i$ is multi-teacher connector logits tailored for the $i^{th}$ task and $ \omega_F^i$ is the adjusted weight determined by Equ.~\ref{equ_omega}. $p_T^i$ represents the logits from $i^{th}$ pretrained teacher, The algorithm is briefed in the paper appendix.

\subsection{Summary of method}
The proposed method facilitates dynamic adjustment of the teacher's knowledge weights (by trainable weight $\omega$) based on the student model's performance on the validation set $\mathbb{D}_{val}$, jointly optimizing them, thereby promoting the development of a more generalized student model. This approach alleviates the learning burden on the student model. Moreover, it underscores the necessity to revisit previously learned objectives by a trajectory knowledge distillation approach when the student model's performance deteriorates and knowledge forgetting occurs. The overall process of the method is summarized in Algorithm~\ref{alg:1}.
\begin{algorithm}[ht]
\caption{The algorithm of our method.}
\label{alg:1}
\begin{algorithmic}[1] 
\REQUIRE ~~\\ 
    $N$ single-task pre-trained teacher models $T_1,\cdots,T_N$ for $N$ tasks. Two datasets $\mathbb{D}_{train}, \mathbb{D}_{val}$ with data numbers $Num_t, Num_v$. Multi-task student model $S$ and feature fusion model $T$.
\ENSURE ~~\\ 
    A small well distilled student model $S$.
    \STATE \textbf{Initialization:} 
    \STATE Randomly initialize $S$ and $T$. Setting $\omega_F^1,\cdots,\omega_F^N=1$, in Eqn (9)) to 1.
    \STATE \textit{\#Training Procedural}
    \FOR{$i \;in \;Num_t$}
        \IF{i=0}
            \STATE Calculating maximum scores $a_{S}^{i,t=0},\cdots,a_{S}^{N,t=0}$ for each task on $\mathbb{D}_{val}$ by $T_1,\cdots,T_N$. 
        \ENDIF
        \STATE Training $F$ using Eqn (9).
        \STATE Update trajectory map by Eqn (7). Delete $J_F^i$ when $len(J_F^i)>t$.
        \STATE Training $S$ using Eqn (4).
        \IF{$i$ == Validation Iteration}
            \STATE \textit{\#Validation Procedural}
            \STATE Calculating student scores by Eqn (1).
            \STATE Update training weight $\omega_T^1,\cdots,\omega_T^N$ by Eqn (2).
        \ENDIF
    \ENDFOR
\end{algorithmic}
\end{algorithm}

\begin{table}[t]
    \centering
    \renewcommand{\arraystretch}{1.18}
    \setlength{\tabcolsep}{1.8mm}{
    \begin{tabular}{c|c|cccc|c}
        \toprule
        {} & \multirow{3}{*}{\textbf{Method}} & \multicolumn{2}{c}{\textbf{Semantic} $\uparrow$} & \multicolumn{2}{c|}{\textbf{Depth} $\downarrow$}  & $\bigtriangleup \uparrow$ \\
        {} & {} & \multirow{2}{*}{mIoU} & Pixel & Abs & Rel & \multirow{2}{*}{$(\%)$}\\
        {} & & & Acc & Err & Err&\\
        \hline
        Teacher1 & Independent & 0.8258 & 0.9565 & -& - & -\\
        Teacher2 & Independent & - & - & 0.0101 & 26.5828 & -\\
        Student & Naive MTL & 0.8149 & 0.9518 & 0.0105 & 36.8269 & 0.00\\
        \hline
        {} & IMTL & 0.7887 & 0.9478 & 0.0204 &45.5143 &-30.38\\
        MTL- & MGDA & 0.7963 & 0.9429 & 0.0270 & 33.4544 &-37.80\\
        Based & CAGrad & 0.8135 & 0.9523 & \textbf{0.0099} & 33.1876 & 3.87\\
        {} & FAMO & 0.8157 & 0.9525 & 0.0109 & 36.3300 & -0.57\\
        \cline{1-7}
         & KDAM & \underline{0.8163} & \underline{0.9526} & 0.0107 & 35.4379 & 0.53\\
        {KD} & AMTL & 0.8132 & 0.9520 & 0.0112 & \underline{27.8423} & 4.34\\
        {Based} & OKD & 0.8146 & 0.9523 & \underline{0.0106} & 34.9421 & 1.05\\
        {} & Ours  & \textbf{0.8173} & \textbf{0.9528} & 0.0114 & \textbf{20.5076} & \textbf{9.04}\\
        \bottomrule
        \end{tabular}}
    \caption{\small Multi-task learning results on Cityscapes dataset. The teachers and the multi-task student are DeeplabV3-Res101 and DeeplabV3-Res50. Each experiment is conducted with the same random seed. The underscore indicates the second-best.}
    \vspace{-5mm}
    \label{tab:citys}
\end{table}

\section{Experiments}
\subsection{Experimental Setup}
\vspace{3pt}\noindent\textbf{Datasets.} We conduct experiments on two benchmark datasets for depth estimation and scene segmentation: \textbf{Cityscapes}~\cite{cordts2016cityscapes} has street scenes from over 50 cities with 2,975 images for training and 500 for testing. \textbf{NYU-v2}~\cite{silberman2012indoor} comprises  1,449 images from 464 scenes of 3 cities, with 654 for testing. By default, 15\% of the training data is reserved as a validation set and utilized 85\% for the train. The preprocessing methods used for the Cityscapes and NYU-v2 datasets are followed~\cite{liu2019end,ye2022inverted}. Moreover, we also conduct experiments on multi-class dataset \textbf{CelebA}~\cite{liu2015deep}, comprising 40 classification tasks.

\vspace{3pt}\noindent\textbf{Evaluation metric.} Following previous work~\cite{liu2019end}, the Semantic task is evaluated with mean Intersection over Union (mIoU) and pixel accuracy (Pix Acc). Absolute error (Abs Err) and relative error (Rel Err) are used in the Depth task. The Normal task is evaluated with mean and median angle distance (Mean and Median). We calculate the overall improvement across multiple tasks using the equation $\bigtriangleup _{MTL}=\frac{1}{N} {\textstyle \sum_{i=1}^{N}} \left( -1 \right)^{l_i} \left ( S_{com}^i-S_{ml}^i \right )/S_{ml}^i\times 100\% $, where $S_{com}^i, S_{ml}^i$ are $i_{th}$ task's scores of comparison methods and uniform multi-task student method, $l_i=1$ if a higher value is better for a criterion $S^i$.

\vspace{3pt}\noindent\textbf{Implementation.} We use DeeplabV3 with ResNet-101~\cite{chen2018encoder} and Segformer~\cite{xie2021segformer} with encoder Mit-B1 for the single-task teacher networks. For the student networks, we use various architectures: DeeplabV3 with ResNet-50 and Segformer with Mit-B0. We train each model with Adam optimizer with a batch size of 8 for the Cityscapes (base learning rate (lr) is 1e-4) and batch size of 2 for NYU-v2 (base lr is 2e-5). For two dataset, the weight decays are set to 1e-4 and 1e-6, and learning rates for the feature fusion module are set to 1e-5 and 2e-6. We use SGD optimizer with a lr of 0.001 and momentum of 0.1 to optimize the training weights $\omega$. Hyperparameters $\lambda, \alpha, k$ are set to 1, 1.5, 10.
%

\vspace{3pt}\noindent\textbf{Baselines} We compare with recent strong baselines of two categories. 1) Advanced MTL methods including \textbf{IMTL}~\cite{liu2021towards},\textbf{MGDA}~\cite{sener2018multi},\textbf{CAGrad}~\cite{liu2021conflict},\textbf{FAMO}~\cite{liu2024famo}; 2) MTL with knowledge distillation strategy including \textbf{KDAM}~\cite{xu2023multi}, \textbf{AMTL}~\cite{yun2023achievement}, and also online distillation method \textbf{OKD}~\cite{jacob2023online}.

\subsection{Comparison Results}
\vspace{3pt}\noindent\textbf{Quantitative results.} We summarized the results in Table~\ref{tab:citys} and Table~\ref{tab:nyu} on Cityscapes and NYU-v2, separately. We mainly have three observations: 1) Although MTL unified the two tasks in a single model, the performance is hardly satisfying. Some advanced MTL method even worsen the performance than a Vanilla-MTL. 2) Overall, distillation improves MTL performance. 3) Our method clearly outperforms the rest on both tasks on most scenarios, with an overall 9.04\% and 2.83\% average task improvement. 

\begin{figure*}[!ht]
    \centering
	\addtocounter{figure}{0} 
	\centering
  \hspace{-9pt}  
	\subfigure{
		\begin{minipage}[t]{0.105\linewidth}
			\includegraphics[width=1\linewidth,height=0.6\linewidth]{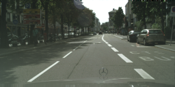}
		\end{minipage}%
	}%
    \hspace{-6pt}
	\subfigure{
		\begin{minipage}[t]{0.105\linewidth}
				\includegraphics[width=1\linewidth,height=0.6\linewidth]{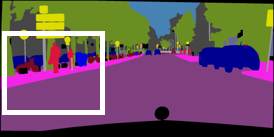}
		\end{minipage}%
	}%
    \hspace{-6pt}
	\subfigure{
		\begin{minipage}[t]{0.105\linewidth}
			\includegraphics[width=1\linewidth,height=0.6\linewidth]{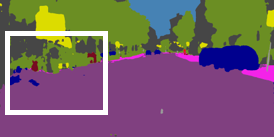}
		\end{minipage}
	}%
    \hspace{-9pt}
	\subfigure{
		\begin{minipage}[t]{0.105\linewidth}
			\includegraphics[width=1\linewidth,height=0.6\linewidth]{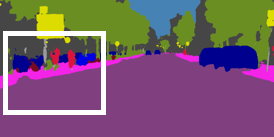}
		\end{minipage}
	}%
    \hspace{-9pt}
	\subfigure{
		\begin{minipage}[t]{0.105\linewidth}
			\includegraphics[width=1\linewidth,height=0.6\linewidth]{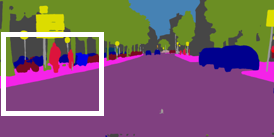}
		\end{minipage}
	}%
    \hspace{-9pt}
	\subfigure{
		\begin{minipage}[t]{0.105\linewidth}
			\includegraphics[width=1\linewidth,height=0.6\linewidth]{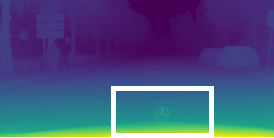}
		\end{minipage}
	}%
   \hspace{-9pt}
	\subfigure{
		\begin{minipage}[t]{0.105\linewidth}
			\includegraphics[width=1\linewidth,height=0.6\linewidth]{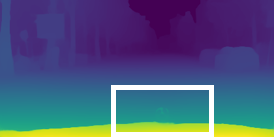}
		\end{minipage}
	}%
   \hspace{-9pt}
	\subfigure{
		\begin{minipage}[t]{0.105\linewidth}
			\includegraphics[width=1\linewidth,height=0.6\linewidth]{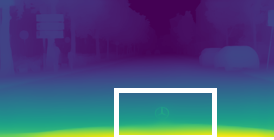}
		\end{minipage}
	}%
  \hspace{-9pt}
	\subfigure{
		\begin{minipage}[t]{0.105\linewidth}
			\includegraphics[width=1\linewidth,height=0.6\linewidth]{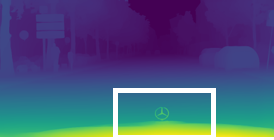}
		\end{minipage}
	}%
  \\
 \hspace{-9pt}
	\subfigure{
		\begin{minipage}[t]{0.105\linewidth}
			\includegraphics[width=1\linewidth,height=0.6\linewidth]{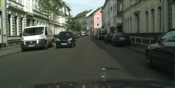}
            \caption*{\small Image}
		\end{minipage}%
	}%
     \hspace{-6pt}
	\subfigure{
		\begin{minipage}[t]{0.105\linewidth}
				\includegraphics[width=1\linewidth,height=0.6\linewidth]{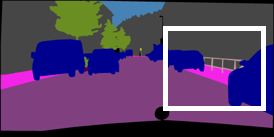}
            \caption*{\small GT/Seg}
		\end{minipage}%
	}%
     \hspace{-6pt}
	\subfigure{
		\begin{minipage}[t]{0.105\linewidth}
			\includegraphics[width=1\linewidth,height=0.6\linewidth]{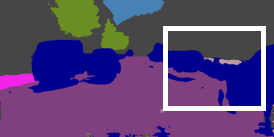}
            \caption*{\small Naive MTL}
		\end{minipage}
	}%
     \hspace{-9pt}
	\subfigure{
		\begin{minipage}[t]{0.105\linewidth}
			\includegraphics[width=1\linewidth,height=0.6\linewidth]{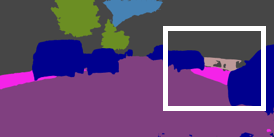}
            \caption*{\small AMTL}
		\end{minipage}
	}%
     \hspace{-9pt}
	\subfigure{
		\begin{minipage}[t]{0.105\linewidth}
			\includegraphics[width=1\linewidth,height=0.6\linewidth]{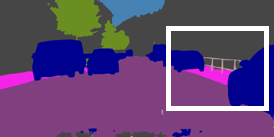}
            \caption*{\small Ours}
		\end{minipage}
	}%
      \hspace{-9pt}
	\subfigure{
		\begin{minipage}[t]{0.105\linewidth}
			\includegraphics[width=1\linewidth,height=0.6\linewidth]{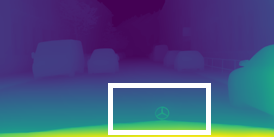}
            \caption*{\small  GT/Depth}
		\end{minipage}
	}%
     \hspace{-9pt}
	\subfigure{
		\begin{minipage}[t]{0.105\linewidth}
			\includegraphics[width=1\linewidth,height=0.6\linewidth]{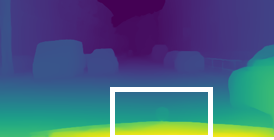}
            \caption*{\small Naive MTL}
		\end{minipage}
	}%
      \hspace{-9pt}
	\subfigure{
		\begin{minipage}[t]{0.105\linewidth}
			\includegraphics[width=1\linewidth,height=0.6\linewidth]{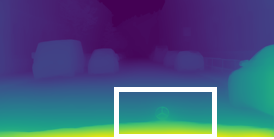}
            \caption*{\small AMTL}
		\end{minipage}
	}%
     \hspace{-9pt}
	\subfigure{
		\begin{minipage}[t]{0.105\linewidth}
			\includegraphics[width=1\linewidth,height=0.6\linewidth]{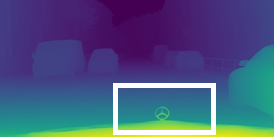}
            \caption*{\small Ours}
		\end{minipage}
	}%
    \vspace{-9pt}
    \caption{\small Visualization examples from Cityscapes for (Column 2-5) semantic segmentation and (Column 6-9) depth estimation.}
    \label{fig:visulization}
    \vspace{-2mm}
\end{figure*}

\begin{table}[t]
    \centering
    \begin{tabular}{c|cc|cc|c}
        \toprule
        \multirow{2}{*}{\textbf{Method}} & \multicolumn{2}{c|}{\textbf{Semantic} $\uparrow$} & \multicolumn{2}{c|}{\textbf{Depth} $\downarrow$} & $\bigtriangleup \uparrow$\\
        {} & mIoU & Pixel Acc & Abs Err & Rel Err & $(\%)$\\
        \hline
        Teacher1 & 0.4448 & 0.7248 & - & - & -\\
        Teacher2 & - & - & 0.1544 & 0.3923 & -\\
        Naive MTL & 0.4332 & 0.7098 & 0.1585 & 0.4053 & 0.00\\
        \hline
        IMTL & 0.4252 & 0.7084 & 0.1563 & 0.4067 & -0.25\\
        MGDA & 0.4272 & 0.7140 & 0.1547 & 0.3997 & 0.75\\
        CAGrad & 0.4313 & 0.7114 & 0.1554 & 0.3984 & 0.86\\
        FAMO & 0.4171 & 0.7053 & \underline{0.1546} & \underline{0.3912} & 0.40\\
        \hline
        KDAM & 0.4279 & 0.7110 & 0.1554 & 0.3990 & 0.63\\
        AMTL & \underline{0.4371} & \underline{0.7172} & 0.1583 & 0.3946 & 1.18\\
        OKD & 0.4351 & 0.7126 & 0.1557 & 0.3928 & 1.42\\
        Ours & \textbf{0.4381} & \textbf{0.7177} & \textbf{0.1518} & \textbf{0.3857} & \textbf{2.83}\\
        \bottomrule
    \end{tabular}
    \caption{\small Results of ResNet backbone teachers and the multi-task student on NYU-v2 dataset. }
    \label{tab:nyu}
\end{table}

\vspace{3pt}\noindent\textbf{Qualitative results.} We also visualize a few examples in Fig~\ref{fig:visulization} to clearly show the improvement on both tasks. Compared to the most strongest baseline AMTL~\cite{yun2023achievement}, we can see that our method produces more accurate
 predictions.

\subsection{Ablation Study and Discussion}
\begin{table}[t]
    \centering
    \renewcommand{\arraystretch}{1.05}
    \begin{tabular}{c|cc|cc}
        \toprule
        \multirow{3}{*}{\textbf{Method}} & \multicolumn{2}{c|}{\textbf{Semantic} $\uparrow$} & \multicolumn{2}{c}{\textbf{Depth} $\downarrow$} \\
        {} & mIoU & Pixel Acc & Abs Err & Rel Err\\
        \hline
        \ding{172}: Naive MTL & 0.8149 & 0.9518 & 0.0105 & 36.8269\\
        \ding{172}$+$Adap-KD & 0.8155 & 0.9526 & 0.0118 & 23.9719\\
        \ding{172}$+$Traj-KD & 0.8169 &0.9532 & 0.0131 & 29.6433\\
        \ding{172}$+$Adap-KD$+$Traj-KD & \textbf{0.8173} & \textbf{0.9528} & \textbf{0.0114} & \textbf{20.5076}\\
        \bottomrule
    \end{tabular}
    \caption{\small Ablation study of our method. Basic is the uniform multi-task student, 'Adap-KD' represents the feedback-based multi-teacher adaptive distillation, and 'Traj-KD' is the trajectory-based knowledge distillation.}
    \label{tab:over_abla}
\end{table}

\vspace{3pt}\noindent\textbf{Overall Ablation Study.} To demonstrate the effectiveness of our approach, we present the results of incrementally integrating our dynamic teacher knowledge distillation (Adap-KD) and trajectory-based knowledge distillation (Traj-KD) methods into the baseline approach. As indicated in Table~\ref{tab:over_abla}, the Adap-KD approach significantly enhances overall performance in the Depth task by effectively leveraging the teacher's knowledge. The distillation of trajectory (Traj-KD) enhances performance in the Semantic task further.

\begin{table}[!t]
    \centering
    \begin{tabular}{c|cc|c|c}
    \toprule
        \multirow{2}{*}{\textbf{Method}} & \multicolumn{2}{c|}{\textbf{Training} $\downarrow$} & \textbf{Semantic} $\uparrow$ & \textbf{Depth} $\downarrow$\\
        {} & \textbf{Memory} & \textbf{Time} & {mIoU} & Rel Err\\
        \hline
        Adap-KD$+$EATA & 10.85 GB & 19.55h & 0.8145 & 24.5847 \\
        Ours & 9.33 GB & 16.87 h & 0.8173 & 20.5076\\
    \bottomrule
    \end{tabular}
    \caption{\small Storage and computation comparison with EATA~\cite{niu2022efficient}.}
    \label{tab:abla_eata}
\end{table}

\vspace{3pt}\noindent\textbf{Computation and storage comparison.} To validate the low-cost feature of our approach, we compare both the memory cost and the training time with EATA~\cite{niu2022efficient}, a popular approach regulated by the Fisher information-based parameter. The setting is based on Deeplabv3+ResNet on Cityscapes. Results in Table~\ref{tab:abla_eata} highlight the superior efficiency of our proposed trajectory distillation module.   


\begin{table}[!t]
    \centering
    \renewcommand{\arraystretch}{1.05}
    \begin{tabular}{c|cc|cc|c}
        \toprule
        \multirow{2}{*}{\textbf{Method}} & \multicolumn{2}{c|}{\textbf{Semantic} $\uparrow$} & \multicolumn{2}{c|}{\textbf{Depth} $\downarrow$} & $\bigtriangleup \uparrow$\\
        {} & {mIoU} & Pixel Acc & Abs Err & Rel Err & $(\%)$\\
        \hline
        Teacher1 & 0.8210 & 0.9539 & - & - & -\\
        Teacher2 & - & - & 0.0130 & 90.8394 & -\\
        Student & 0.8017 & 0.9487 & 0.0139 & 104.6983 & 0.00\\
        \hline
        IMTL & 0.8005 & 0.9479 & 0.0146 & 57.0608 & 10.06\\
        MGDA & 0.7989 & 0.9479 & 0.0176 & 133.9214 & -13.75\\
        CAGrad & 0.8050 & \underline{0.9493} & 0.0165 & 136.9348 & -12.26\\
        FAMO & 0.8007 & 0.9482 & \underline{0.0139} & \underline{50.4966} & 12.90\\
        \hline
        KDAM & 0.7950 & 0.9454 & 0.0149 & 98.9014 & -0.71\\
        AMTL & 0.8049 & 0.9493 & 0.0146 & 71.9061 & 6.69\\
        OKD & \underline{0.8052} & 0.9492 & 0.0147 & 61.4731 & 9.00\\
        Ours & \textbf{0.8072} & \textbf{0.9498} & \textbf{0.0137} & \textbf{59.7787} & \textbf{11.29}\\
        \bottomrule
    \end{tabular}
    \caption{\small Segformer results on Cityscapes dataset. The encoders of teachers and the student are Mit-B1 and Mit-B0.}
    \label{tab:seg_city}
\end{table}


\vspace{3pt}\noindent\textbf{Effects of model backbones} We also replace the backbone of the teacher and student models with Segformer~\cite{xie2021segformer} architectures. Results in Table~\ref{tab:seg_city} show that we can robustly achieve similar improvements on both architectures.

\vspace{3pt}\noindent\textbf{Effects of hyperparameter $\alpha$ and $\beta$.} We analyzed the impact of different values of $\alpha$ and $\beta$ in Eqn.~\ref{equ_omega} and found that when $\alpha$ is too large or too small, the dynamic weight $\omega$ will updates slowly, leading to suboptimal performance. The best results were achieved when $\alpha$ was within the range of [1, 2] and $\beta$ was 0.001 (Figure~\ref{fig:abla_a}).

\vspace{3pt}\noindent\textbf{Enhancement with multi-modal knowledge.} The image-encoder of our MTL model could be used alone for both tasks, or as an image encoder that could be combined with other modalities (e.g., texts) for enhancement. Following~\cite{zhou2022learning}, our image feature is combined with the text feature from a textual prompt by a frozen CLIP text encoder via scaled dot-product-based attention~\cite{vaswani2017attention}. We observe a multi-modality enhancement in Table~\ref{tab:multi-modal} with this extremely simple setting.  We believe further improvement could be achieved with advanced fusion and we left the exploration for future. 

\vspace{3pt}\noindent\textbf{Scalability to 40 tasks} To extend our methods to a wider range of multi-task scenarios on the CelebA dataset. We employ a single multi-task teacher model based on the MTAN~\cite{liu2019end} architecture, which extends multiple branches from the backbone network (ResNet-101). Then the student model (DeeplabV3 + ResNet-50) learns multi-source knowledge through the feature fusion module ($F(\cdot)$ in Section~\ref{sec: overview}), which processes the features of all tasks. The mean accuracy of 40 tasks demonstrates that our method achieves superior performance.

\begin{table}[!t]
    \centering
    \begin{tabular}{c|cc|cc}
    \toprule
        \multirow{2}{*}{\textbf{Method}} & \multicolumn{2}{c|}{\textbf{Semantic} $\uparrow$} & \multicolumn{2}{c}{\textbf{Depth} $\downarrow$}\\
        {} & {mIoU} & Pixel Acc & Abs Err & Rel Err\\
        \hline
         Ours $+$ w/o Text Prompt & 0.8072 & 0.9498 & 0.0137 & 59.7787\\
         Ours $+$ w/ Text Prompt & 0.8075 & 0.9499 & 0.0135 & 55.2386\\
    \bottomrule
    \end{tabular}
    \caption{\small A simple multimodality enhancement with a trainable text prompt. The setting is with Mit-B1 and Mit-B0 encoder on Cityscapes.}
    \vspace{-3mm}
    \label{tab:multi-modal}
\end{table}

\section{Conclusion}
\noindent In this work, we propose an adaptive multi-task knowledge distillation method for joint scene segmentation and depth prediction. We first periodically adjust the knowledge imparted by each task-specific teacher based on the MTL-student's performance feedback on unseen data. Additionally, we introduce a lightweight trajectory distillation to address knowledge forgetting which commonly exists in multi-task settings with much lower storage and computation lost. The results multi-task student model could be used alone for both tasks, or be used as an image encoder and combined with other modalities for a multi-modal enhancement. Compared to the state-of-the-art methods, we achieve a clear improvement on the transportation datasets including Cityscapes and NYU-v2. 

\textbf{Acknowledgments.} This work was supported by the National Key R\&D Program of China (No. 2024YFB4505502), and the National Natural Science Foundation of China (No. 62432007, 62272390).

\begin{figure}[!t]
    \addtocounter{figure}{0} 
    \centering
	\subfigure{
		\begin{minipage}[t]{0.48\linewidth}
			\includegraphics[width=1\linewidth,height=0.6\linewidth]{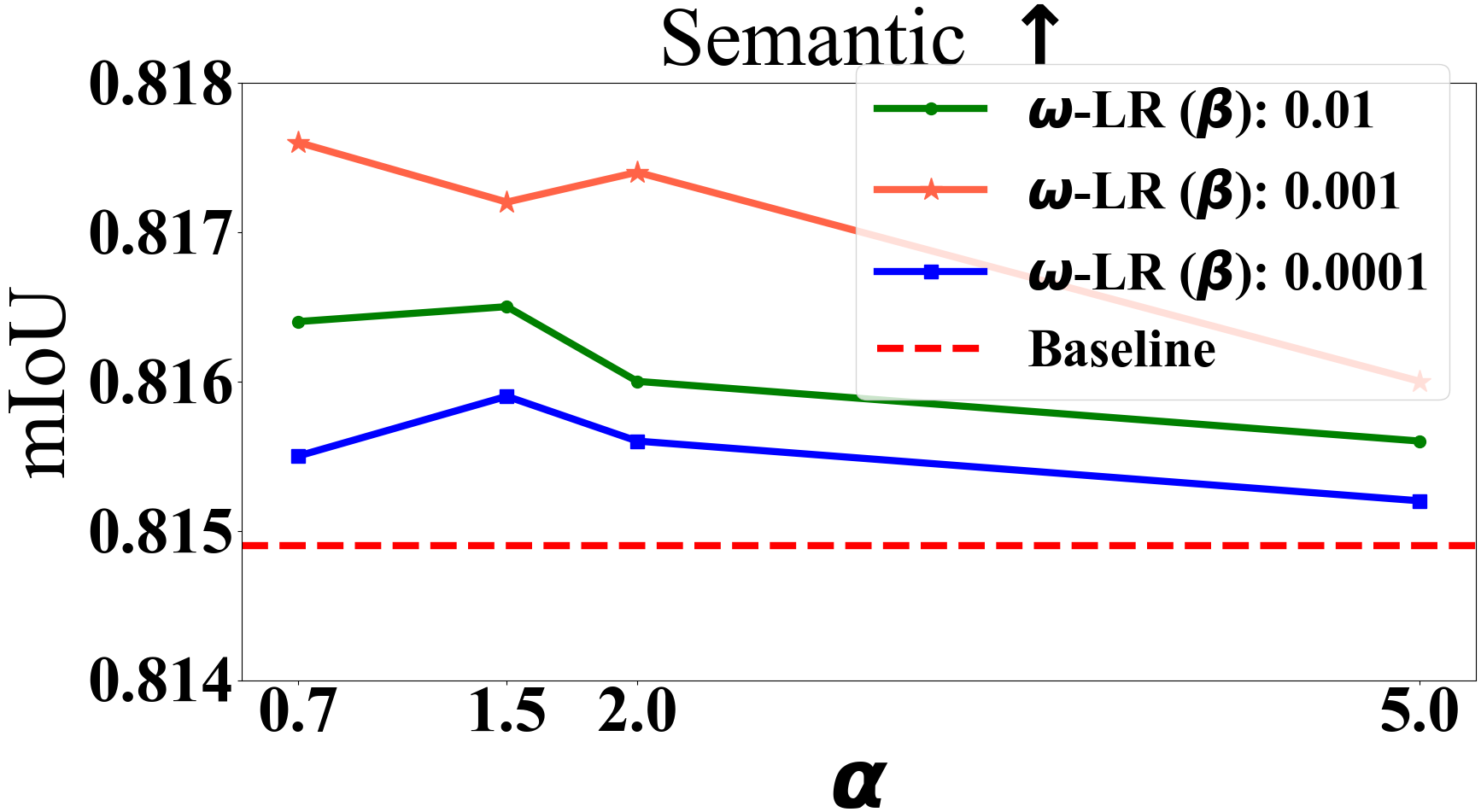}
		\end{minipage}%
	}%
	\subfigure{
		\begin{minipage}[t]{0.48\linewidth}
				\includegraphics[width=1\linewidth,height=0.6\linewidth]{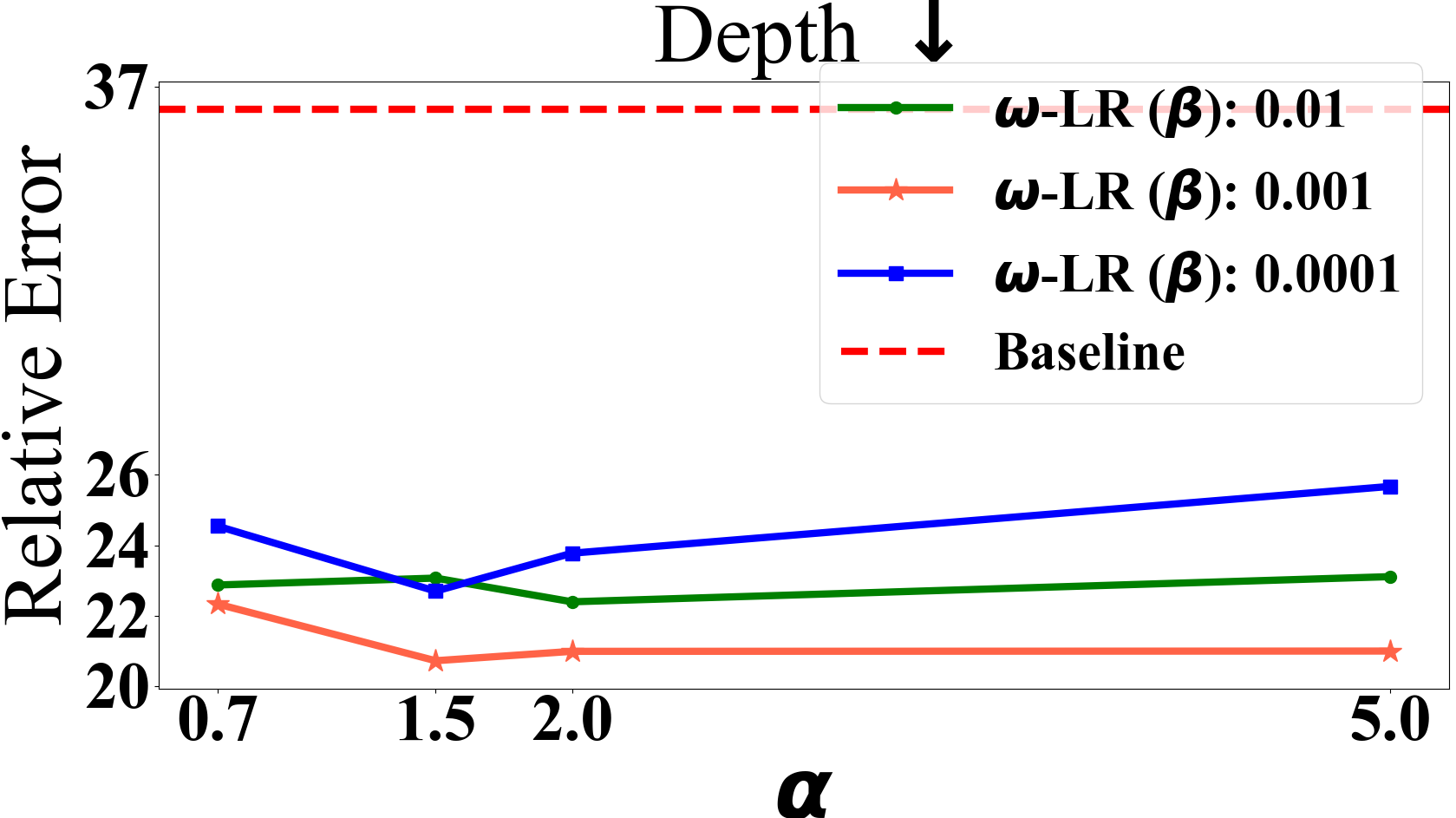}
		\end{minipage}%
	}%
    \vspace{-3mm}
    \caption{\small Ablation results of different $\alpha$ and different optimization learning rates $\beta$ (Eqn.~\ref{equ_omega}) when updates $\omega$.}
    \label{fig:abla_a}
\end{figure}

\begin{figure}
    \centering
    \includegraphics[width=3.in]{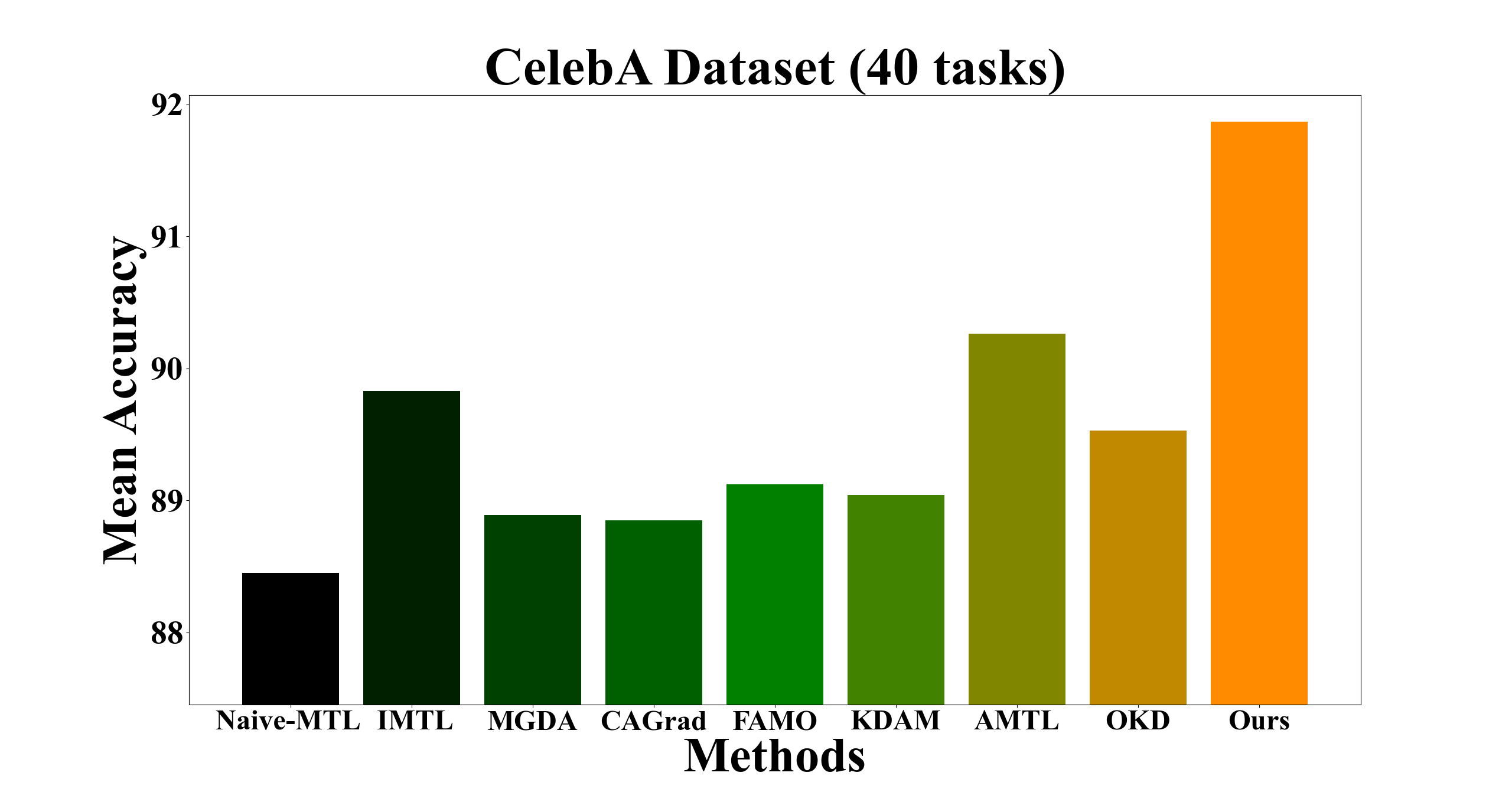}
    \caption{Illustration of further enhancement of the method by multi-modal knowledge.}
    \label{fig:multi-modal}
\end{figure}

\bibliographystyle{IEEEbib}
\bibliography{icme}

\end{document}